\documentclass[11pt,a4paper]{article}
\usepackage[hyperref]{emnlp-ijcnlp-2019}
\usepackage{times}
\usepackage{latexsym}
\usepackage{url}

\usepackage{todos}
\usepackage{bm}
\usepackage{color}
\usepackage{multirow}
\usepackage{amsmath}
\usepackage{subfig}
\usepackage{enumitem}
\usepackage{amsfonts}
\usepackage{multirow}
\usepackage{makecell}
\usepackage{arydshln}

\usepackage{multicol}

\usepackage{graphicx} 
\usepackage{graphics}

\usepackage{balance}

\definecolor{zptu}{RGB}{18, 141, 21}

\definecolor{electricviolet}{rgb}{0.56, 0.0, 1.0}

\aclfinalcopy 

\title{Towards Understanding Neural Machine Translation \\ with Word Importance}

\author{
Shilin He$^{1,2}$   ~~   Zhaopeng Tu$^3$\thanks{~~Zhaopeng Tu is the corresponding author. Work was mainly done when Shilin He was interning at Tencent AI Lab.} ~~  Xing Wang$^3$   ~~  Longyue Wang$^3$  ~~ Michael R. Lyu$^{1,2}$ ~~  Shuming Shi$^3$\\
{$^1$Department of Computer Science and Engineering, The Chinese University of Hong Kong} \\
{$^2$Shenzhen Research Institute, The Chinese University of Hong Kong} \\
$^{1,2}${\tt  \{slhe,lyu\}@cse.cuhk.edu.hk} \\
{$^3$Tencent AI Lab}    \\
$^3${\tt \{zptu,brightxwang,vinnylywang,shumingshi\}@tencent.com}
}

\date{}

\begin{document}
\maketitle
\begin{abstract}


Although neural machine translation (NMT) has advanced the state-of-the-art on various language pairs, the interpretability of NMT remains unsatisfactory. In this work, we propose to address this gap by focusing on understanding the input-output behavior of NMT models. Specifically, we measure the word importance by attributing the NMT output to every input word through a gradient-based method. We validate the approach on a couple of perturbation operations, language pairs, and model architectures, demonstrating its superiority on identifying input words with higher influence on translation performance. Encouragingly, the calculated importance can serve as indicators of input words that are under-translated by NMT models. Furthermore, our analysis reveals that words of certain syntactic categories have higher importance while the categories vary across language pairs, which can inspire better design principles of NMT architectures for multi-lingual translation.

\end{abstract}


\section{Introduction}
\label{sec:introduction}

Neural machine translation (NMT) has achieved the state-of-the-art results on a mass of language pairs with varying structural differences, such as English-French~\cite{bahdanau2014neural,vaswani2017attention} and Chinese-English~\cite{hassan2018achieving}. However, so far not much is known about how and why NMT works, which pose great challenges for debugging NMT models and designing optimal architectures.

The understanding of NMT models has been approached primarily from two complementary perspectives. The first thread of work aims to understand the importance of representations by analyzing the linguistic information embedded in representation vectors~\cite{Shi:2016:EMNLP,Belinkov:2017:ACL} or hidden units~\cite{bau2018identifying,ding2017visualizing}.
Another direction focuses on understanding the importance of input words by interpreting the input-output behavior of NMT models. 
Previous work~\cite{alvarez2017causal} treats NMT models as black-boxes and provides explanations that closely resemble the attention scores in NMT models. However, recent studies reveal that attention does not provide meaningful explanations since the relationship between attention scores and model output is unclear~\cite{Jain2019AttentionIN}.

In this paper, we focus on the second thread and try to open the black-box by exploiting the gradients in NMT generation, which aims to estimate the word importance better. 
Specifically, we employ the {\em integrated gradients} method~\cite{sundararajan2017axiomatic} to attribute the output to the input words with the integration of first-order derivatives. We justify the gradient-based approach via quantitative comparison with black-box methods on a couple of perturbation operations, several language pairs, and two representative model architectures, demonstrating its superiority on estimating word importance. 

We analyze the linguistic behaviors of words with the importance and show its potential to improve NMT models. First, we leverage the word importance to identify input words that are under-translated by NMT models. Experimental results show that the gradient-based approach outperforms both the best black-box method and other comparative methods. Second, we analyze the linguistic roles of identified important words, and find that words of certain syntactic categories have higher importance while the categories vary across language. 
For example, nouns are more important for Chinese$\Rightarrow$English translation, while prepositions are more important for English-French and -Japanese translation.
This finding can inspire better design principles of NMT architectures for different language pairs. For instance, a better architecture for a given language pair should consider its own language characteristics.

\paragraph{Contributions} Our main contributions are:
\begin{itemize}
    \item Our study demonstrates the necessity and effectiveness of exploiting the intermediate gradients for estimating word importance.
    \item We find that word importance is useful for understanding NMT by identifying under-translated words.
    \item We provide empirical support for the design principle of NMT architectures: essential inductive bias (e.g., language characteristics) should be considered for model design.
\end{itemize}


\section{Related Work}


\paragraph{Interpreting Seq2Seq Models}
Interpretability of Seq2Seq models has recently been explored mainly from two perspectives: interpreting  internal representations and understanding  input-output behaviors. Most of the existing work focus on the former thread, which analyzes the linguistic information embeded in the learned representations~\cite{Shi:2016:EMNLP,Belinkov:2017:ACL,Yang:2019:ACL} or the hidden units~\cite{ding2017visualizing,bau2018identifying}. 
Several researchers turn to expose systematic differences between human and NMT translations~\cite{Laubli:2018:EMNLP,schwarzenberg2019train}, indicating the linguistic properties worthy of investigating.
However, the learned representations may depend on the model implementation, which potentially limit the applicability of these methods to a broader range of model architectures. 
Accordingly, we focus on understanding the input-output behaviors, and validate on different architectures to demonstrate the universality of our findings.

Concerning interpreting the input-output behavior, previous work generally treats Seq2Seq models as black-boxes~\cite{li2016understanding,alvarez2017causal}. For example,~\newcite{alvarez2017causal} measure the relevance between two input-output tokens by perturbing the input sequence. However, they do not exploit any intermediate information such as gradients, and the relevance score only resembles attention scores. Recently,~\newcite{Jain2019AttentionIN} show that attention scores are in weak correlation with the feature importance. Starting from this observation, we exploit the intermediate gradients to better estimate word importance, which consistently outperforms its attention counterpart across model architectures and language pairs.

\paragraph{Exploiting Gradients for Model Interpretation}
The intermediate gradients have proven to be useful in interpreting deep learning models, such as NLP models~\cite{mudrakarta2018did,dhamdhere2018important} and computer vision models~\cite{selvaraju2017grad,sundararajan2017axiomatic}. 
Among all gradient-based approaches, the integrated gradients~\citep[IG,][]{sundararajan2017axiomatic} is appealing since it does not need any instrumentation of the architecture and can be computed easily by calling gradient operations. 
In this work, we employ the IG method to interpret NMT models and reveal several interesting findings, which can potentially help debug NMT models and design better architectures for specific language pairs.

\section{Approach}

\subsection{Neural Machine Translation}
In machine translation task, a NMT model $F$: $\textbf{x} \rightarrow \textbf{y}$ maximizes the probability of a target sequence $\textbf{y} = \{y_1,...,y_N\}$ given a source sentence $\textbf{x} = \{x_1,...,x_M\}$:
\begin{equation*}
    P(\textbf{y}|\textbf{x};\bm{\theta}) = \displaystyle \prod_{n=1}^{N} P(y_n|\textbf{y}_{<n},\textbf{x};\bm{\theta})
\end{equation*}
where $\bm{\theta}$ is the model parameter and $\textbf{y}_{<n}$ is a partial translation. At each time step \textit{n}, the model generates an output word of the highest probability based on the source sentence $\textbf{x}$ and the partial translation $\textbf{y}_{<n}$. The training objective is to minimize the negative log-likelihood loss on the training corpus. During the inference, beam search is employed to decode a more optimal translation.
In this study, we investigate the contribution of each input word $x_m$ to the translated sentence ${\bf y}$.

\subsection{Word Importance}
\label{subsec:word-importance}
In this work, the notion of ``word importance'' is employed to quantify the contribution that a word in the input sentence makes to the NMT generations.
We categorize the methods of word importance estimation into two types: {\em black-box} methods without the knowledge of the model and {\em white-box} methods that have access to the model internal information (e.g., parameters and gradients). Previous studies mostly fall into the former type, and in this study, we investigate several representative black-box methods:
\begin{itemize}
    \item{\em Content Words}: In linguistics, all words can be categorized as either content or content-free words. Content words consist mostly of nouns, verbs, and adjectives, which carry descriptive meanings of the sentence and thereby are often considered as important. 
    \item{\em Frequent Words}: We rank the relative importance of input words according to their frequency in the training corpus. We do not consider the top 50 most frequent words since they are mostly punctuation and stop words.
    \item {\em Causal Model}~\cite{alvarez2017causal}: Since the causal model is complicated to implement and its scores closely resemble attention scores in NMT models. In this study, we use {\em Attention} scores to simulate the causal model.
\end{itemize}

Our approach belongs to the white-box category by exploiting the intermediate gradients, which will be described in the next section.

\subsection{Integrated Gradients}

In this work, we resort to a gradient-based method, integrated gradients~\cite{sundararajan2017axiomatic} (IG), which was originally proposed to attribute the model predictions to input features. It exploits the handy model gradient information by integrating first-order derivatives. IG is implementation invariant and does not require neural models to be differentiable or smooth, thereby is suitable for complex neural networks like Transformer. In this work, we use IG to estimate the word importance in an input sentence precisely.

\begin{figure}[t]
    \centering
    \includegraphics[width=0.45\textwidth]{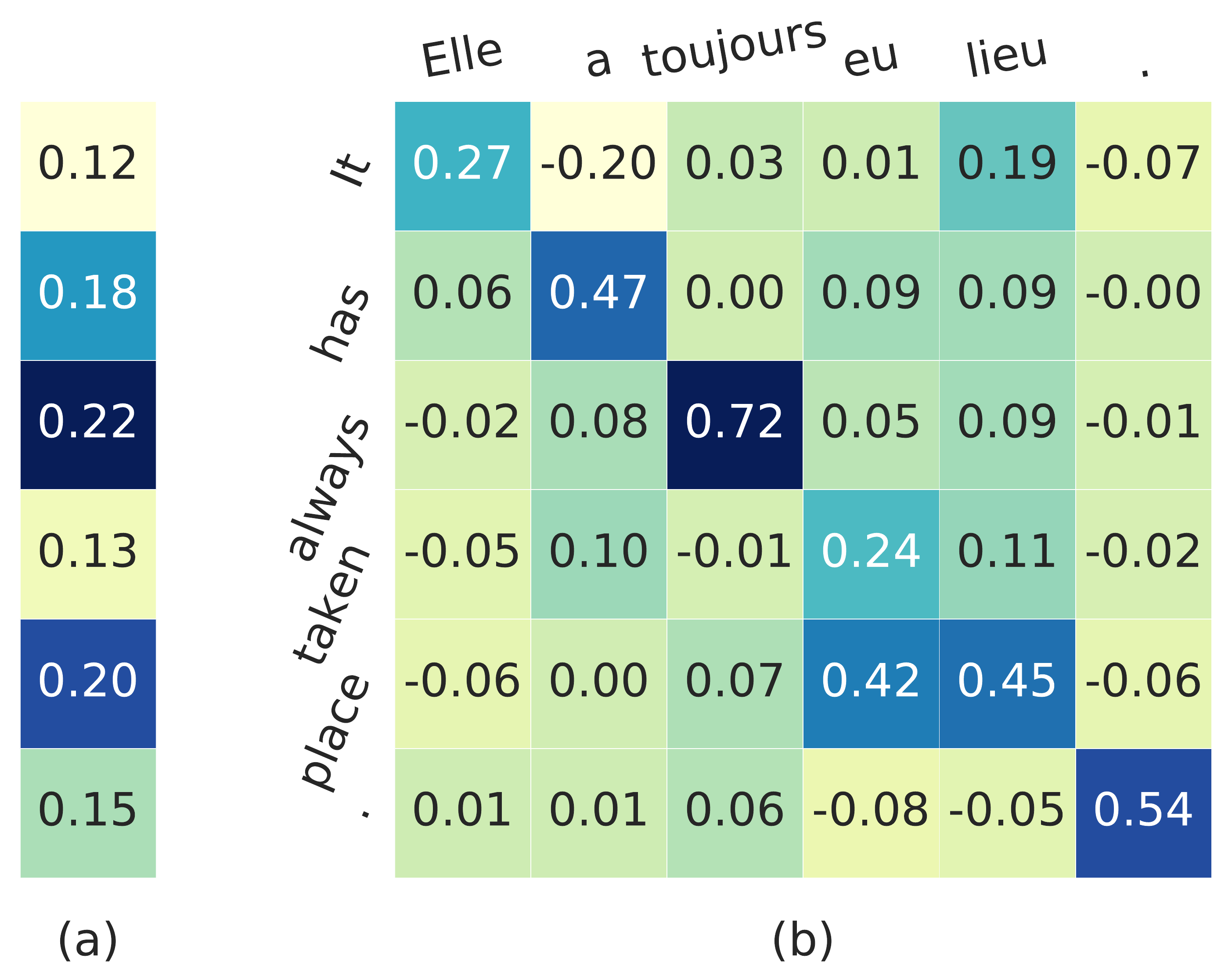}
    \caption{An example of (a) word importance and (b) contribution matrix calculated by \textit{Attribution} (integrated gradients) on English$\Rightarrow$French translation task. Input in English: ``It has always taken place .'' Output in French: ``Elle a toujours eu lieu .''}
    \label{fig:attribution}
\end{figure}

Formally, let $\textbf{x}  = (x_1, ..., x_M)$ be the input sentence and $\textbf{x}'$ be a baseline input. $F$ is a well-trained NMT model, and $F(\textbf{x})_n$ is the model output (i.e., $P(y_n|\textbf{y}_{<n},\textbf{x})$) at time step $n$.  Integrated gradients is then defined as the integral of gradients along the straightline path from the baseline $\textbf{x}'$ to the input $\textbf{x}$. 
In detail, the contribution of the $m^{th}$ word in $\textbf{x}$ to the prediction of $F(\textbf{x})_n$ is defined as follows.
\begin{equation*}\small
\label{eq:IG}
    IG_m^n(\textbf{x}) = (\textbf{x}_m-\textbf{x}'_m) \int_{\alpha=0}^1  \frac{\partial{F(\textbf{x}'+ \alpha(\textbf{x}-\textbf{x}'))_n}}{\partial{\textbf{x}_m}} \mathrm{d}\alpha
\end{equation*}
where $\frac{\partial{F(\textbf{x})_n}}{\partial{\textbf{x}_m}}$ is the gradient of $F(\textbf{x})_n$ w.r.t. the embedding of the $m^{th}$ word. In this paper, as suggested, the baseline input $\textbf{x}'$ is set as a sequence of zero embeddings that has the same sequence length $M$. In this way, we can compute the contribution of a specific input word to a designated output word. Since the above formula is intractable for deep neural models, we approximate it by summing the gradients along a multi-step path from baseline $\textbf{x}'$ to the input \textbf{x}. 
\begin{equation*}\small
\label{eq:IG_appro}
    IG_m^{n}(\textbf{x}) = \frac{(\textbf{x}_m-\textbf{x}'_m)}{S} \sum_{k=0}^{S}  \frac{\partial{F(\textbf{x}'+ \frac{k}{S}(\textbf{x}-\textbf{x}'))_n}}{\partial{\textbf{x}_m}} 
\end{equation*}
where $S$ denotes the number of steps that are uniformly distributed along the path. The IG will be more accurate if a larger S is used. In our preliminary experiments, we varied the steps and found 300 steps yielding fairly good performance. 

Following the formula, we can calculate the contribution of every input word makes to every output word, forming a contribution matrix of size $M \times N$, where $N$ is the output sentence length. Given the contribution matrix, we can obtain the \textit{word importance} of each input word to the entire output sentence. To this end, for each input word, we first aggregate its contribution values to all output words by the \textit{sum} operation, and then normalize all sums through the \textit{Softmax} function. Figure~\ref{fig:attribution} illustrates an example of the calculated word importance and the contribution matrix, where an English sentence is translated into a French sentence using the Transformer model. A negative contribution value indicates that the input word has negative effects on the output word.


\section{Experiment}

\paragraph{Data} To make the conclusion convincing, we first choose two large-scale datasets that are publicly available, i.e., Chinese-English and  English-French. Since English, French, and Chinese all belong to the subject-verb-object (SVO) family, we choose another very different subject-object-verb (SOV) language, Japanese, which might bring some interesting linguistic behaviors in English-Japanese translation. 

For Chinese-English task, we use WMT17 Chinese-English dataset that consists of $20.6$M sentence pairs. 
For English-French task, we use WMT14 English-French dataset that comprises $35.5$M sentence pairs. 
For English-Japanese task, we follow~\citep{morishita2017ntt} to use the first two sections of WAT17 English-Japanese dataset that consists of $1.9$M sentence pairs. Following the standard NMT procedure, we adopt the standard byte pair encoding (BPE)~\cite{sennrich2015neural} with 32K merge operations for all language pairs. We believe that these datasets are large enough to confirm the rationality and validity of our experimental analyses.


\paragraph{Implementation} 
We choose the state-of-the-art Transformer~\cite{vaswani2017attention} model and the conventional RNN-Search model~\cite{bahdanau2014neural} as our test bed. We implement the \textit{Attribution} method based on the Fairseq-py~\cite{gehring2017convolutional} framework for the above models. All models are trained on the training corpus for 100k steps under the standard settings, which achieve comparable translation results. All the following experiments are conducted on the test dataset, and we estimate the input word importance using the model generated hypotheses.

In the following experiments, we compare IG (\textit{Attribution}) with several black-box methods (i.e., \textit{Content}, \textit{Frequency}, \textit{Attention}) as introduced in Section \ref{subsec:word-importance}. In Section~\ref{sec-exp-perturbation}, to ensure that the translation performance decrease attributes to the selected words instead of the perturbation operations, we randomly select the same number of words to perturb (\textit{Random}), which serves as a baseline. 
Since there is no ranking for content words, we randomly select a set of content words as important words. To avoid the potential bias introduced by randomness (i.e., {\em Random} and {\em Content}), we repeat the experiments for 10 times and report the averaged results. 
We calculate the \textit{Attention} importance in a similar manner as the \textit{Attribution}, except that the attention scores use a {\em max} operation due to the better performance.

\paragraph{Evaluation}
We evaluate the effectiveness of estimating word importance by the translation performance decrease. More specifically, unlike the usual way, we measure the decrease of translation performance when perturbing a set of important words that are of top-most word importance in a sentence. The more translation performance degrades, the more important the word is.

We use the standard BLEU score as the evaluation metric for translation performance.  To make the conclusion more convincing, we conduct experiments on different types of synthetic perturbations (Section~\ref{sec-exp-perturbation}), as well as different NMT architectures and language pairs (Section~\ref{sec-exp-models}). 
In addition, we compare with a supervised erasure method, which requires ground-truth translations for scoring word importance (Section~\ref{sec-erasure}).

\begin{figure*}[ht]
    \centering
    \subfloat[\bf Deletion]{
    \includegraphics[width=0.3\textwidth]{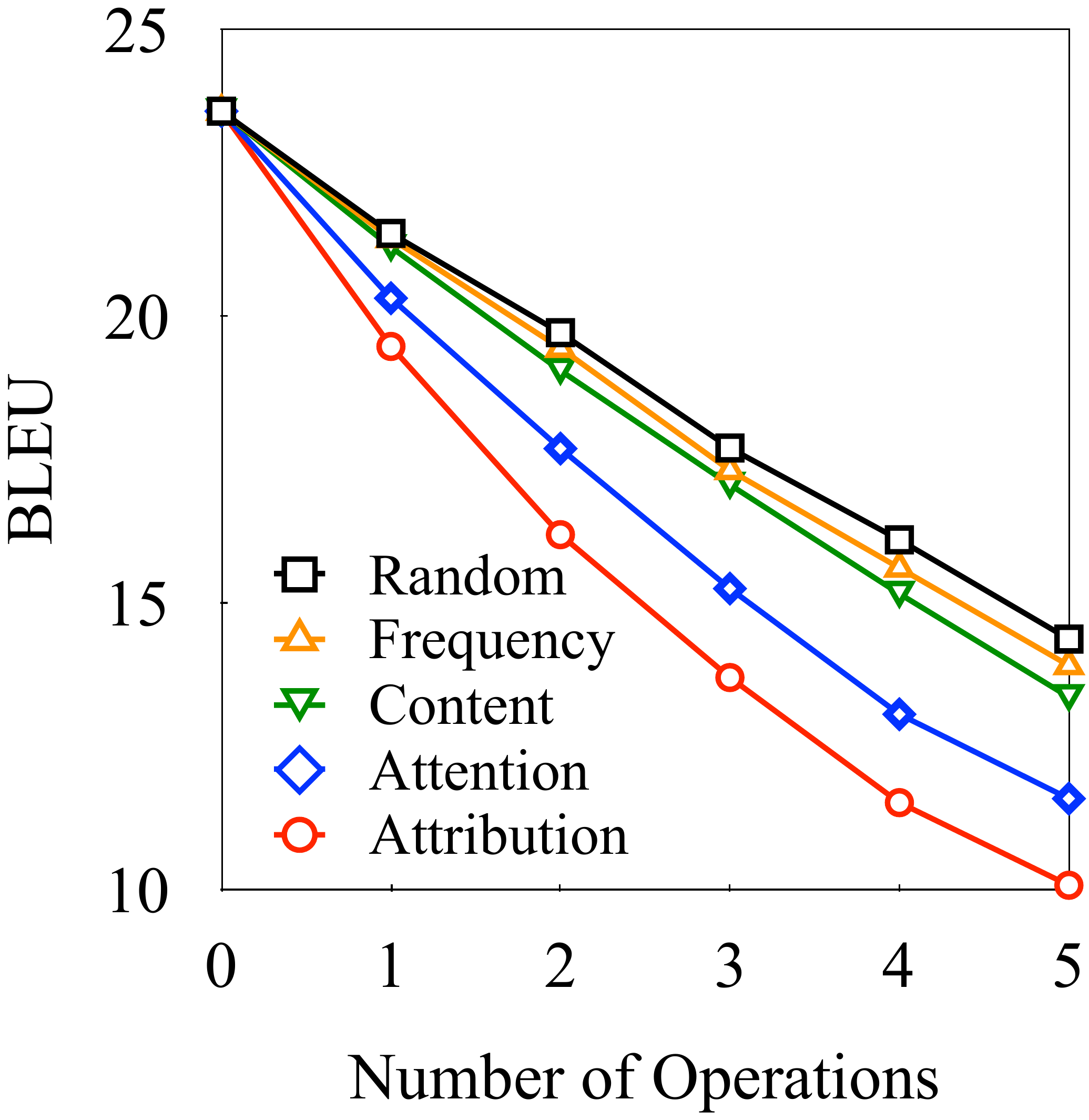}}  \hspace{0.05\columnwidth}
    \subfloat[\bf Mask]{
    \includegraphics[width=0.3\textwidth]{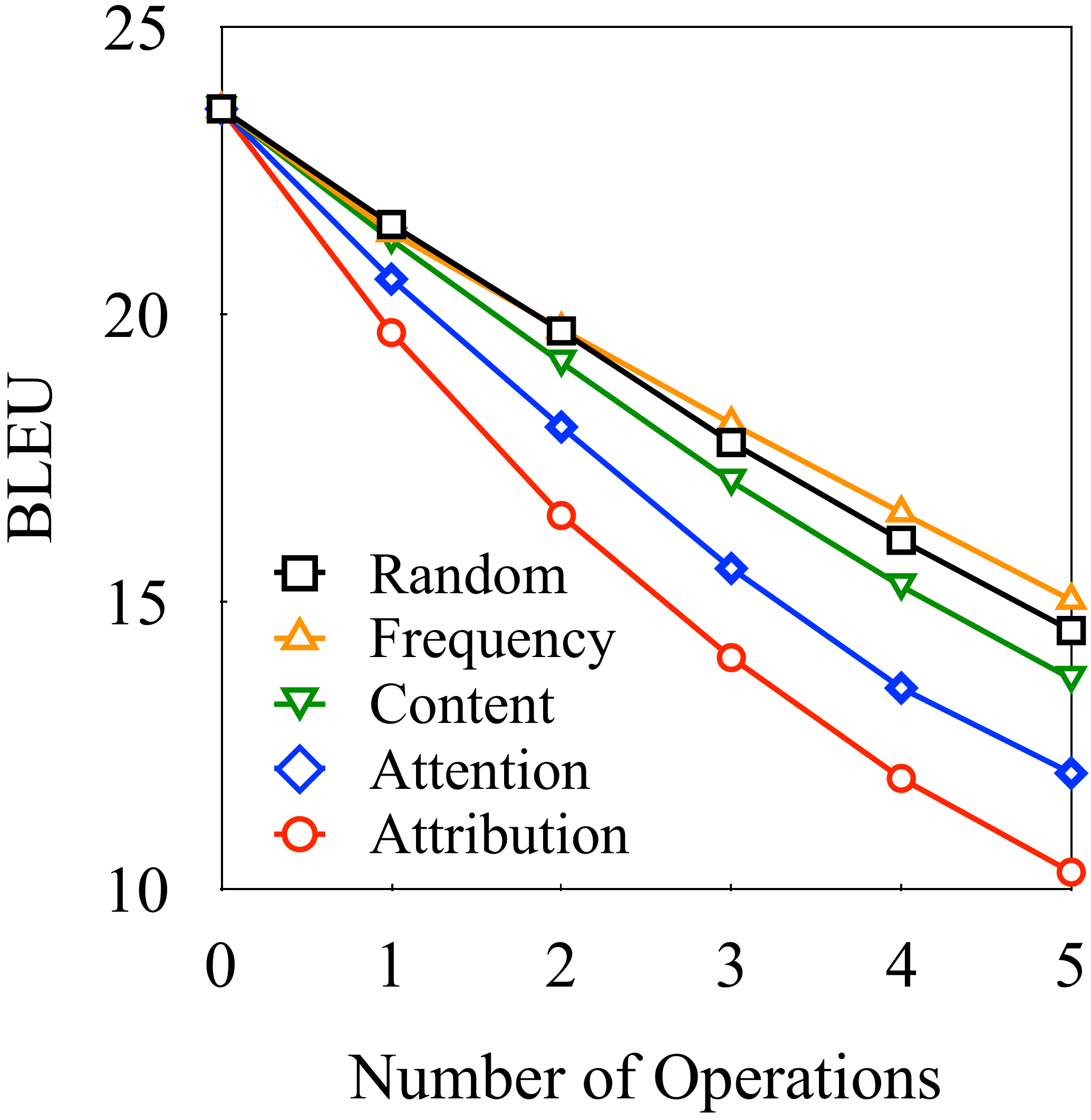}} \hspace{0.05\columnwidth}
    \subfloat[\bf Grammatical Replacement]{
    \includegraphics[width=0.3\textwidth]{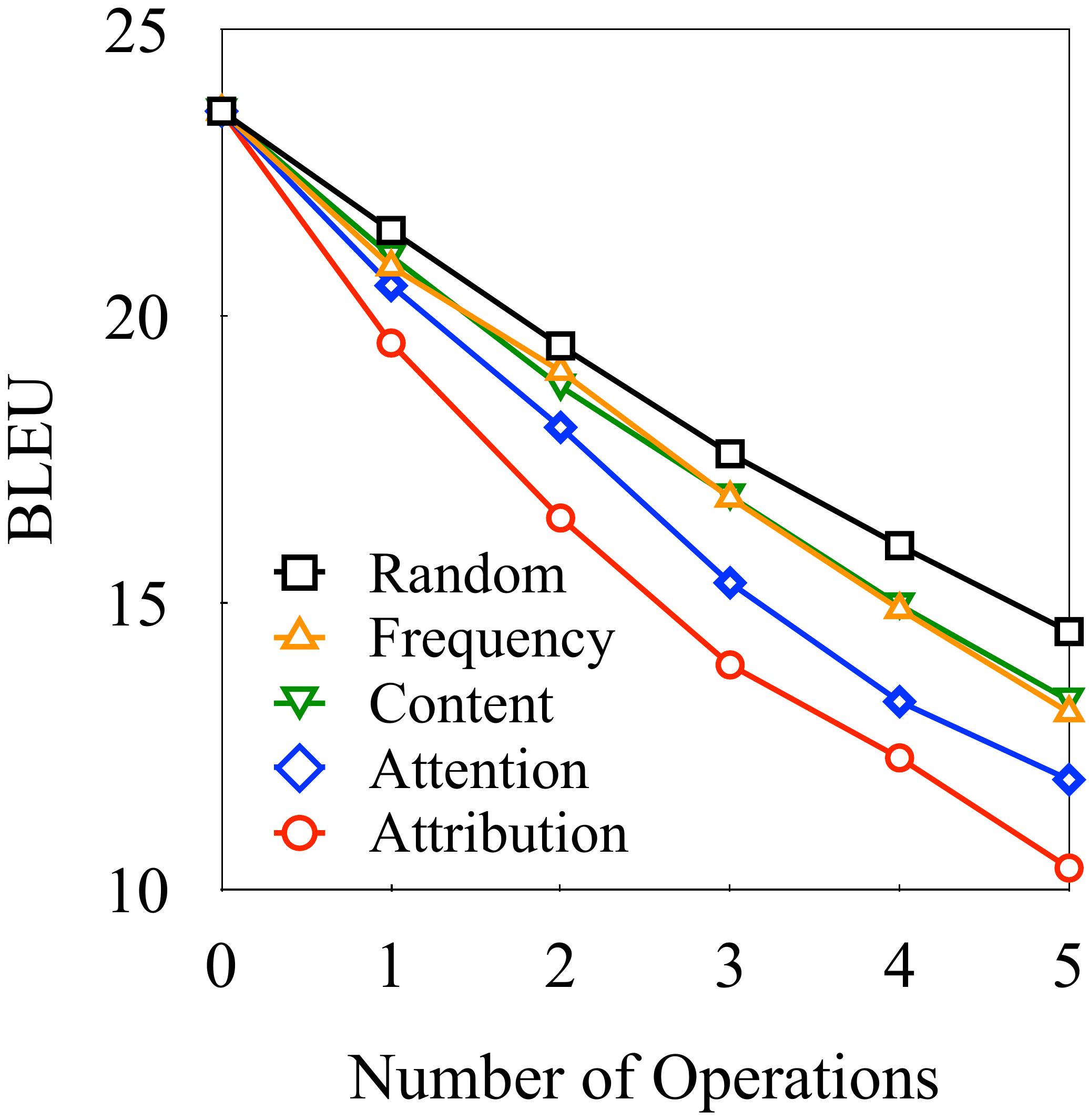}}
\caption{Effect of three types of synthetic perturbations on Chinese$\Rightarrow$English translation using the Transformer.}
\label{fig:perturbations}
\end{figure*}

\begin{figure*}[th]
    \centering
    \subfloat[\bf RNN-Search Model]{
    \includegraphics[width=0.3\textwidth]{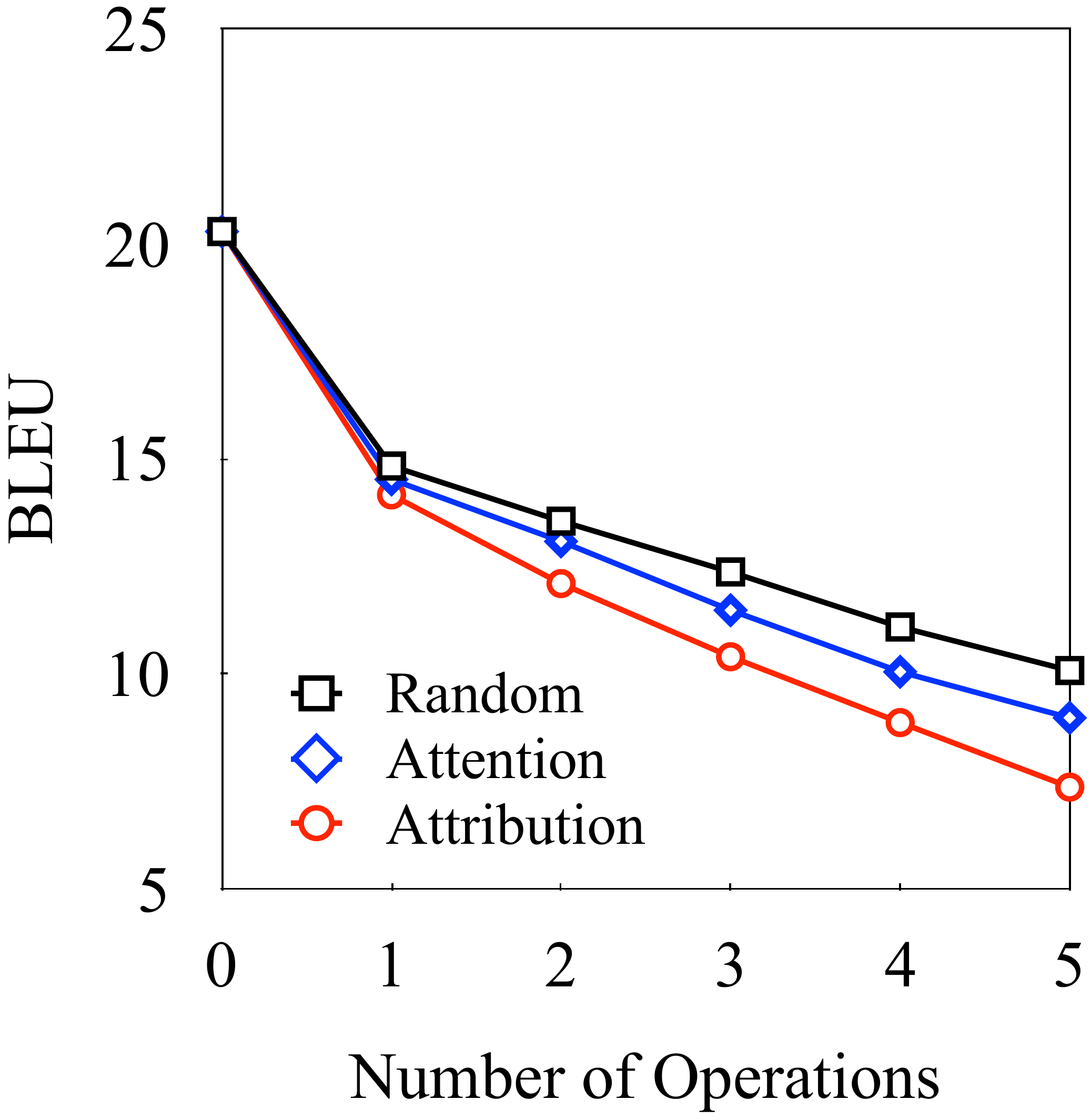}} 
    \hspace{0.05\columnwidth}
    \subfloat[\bf English$\Rightarrow$French]{
    \includegraphics[width=0.3\textwidth]{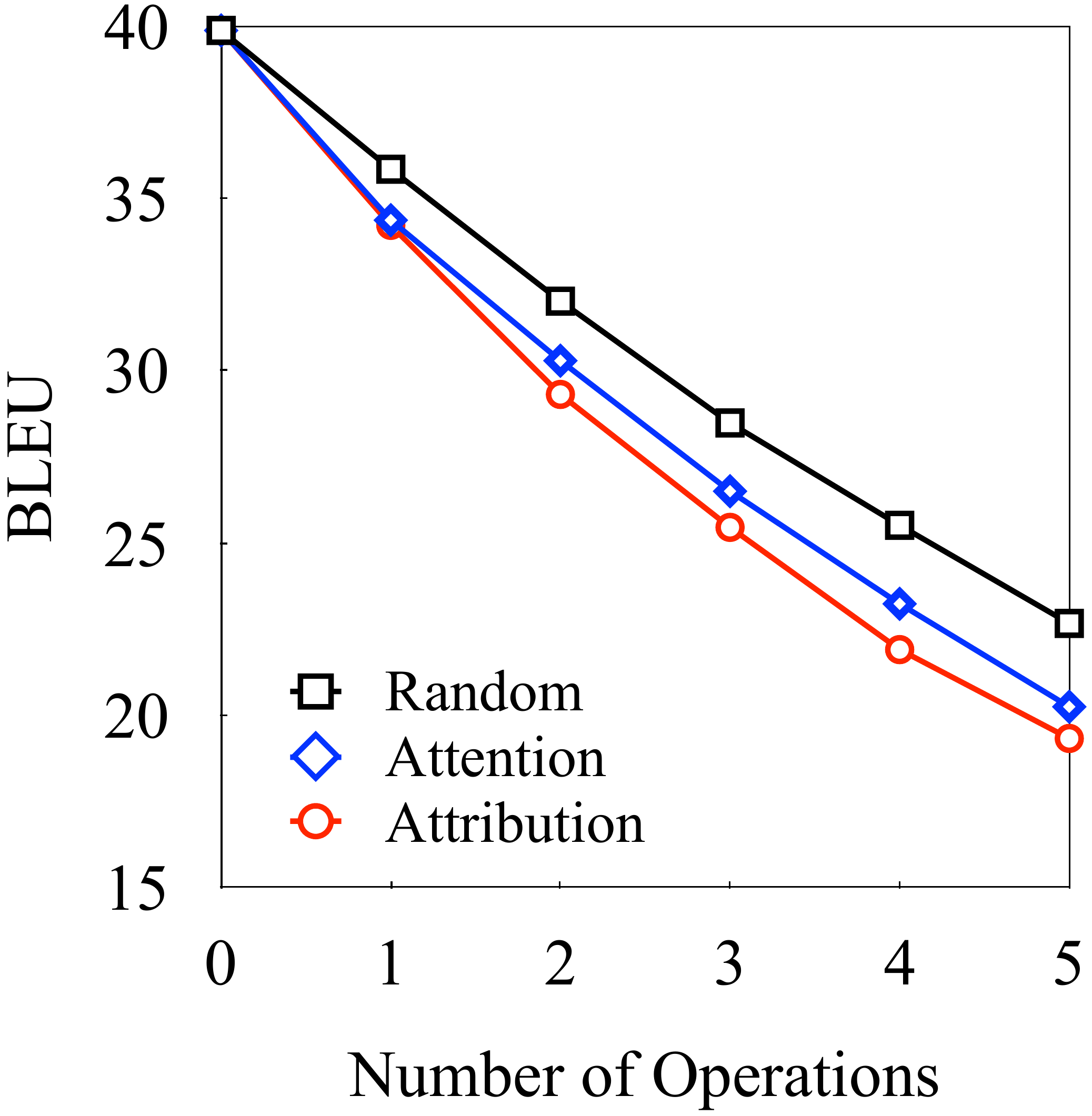}} \hspace{0.05\columnwidth}
    \subfloat[\bf English$\Rightarrow$Japanese]{
    \includegraphics[width=0.3\textwidth]{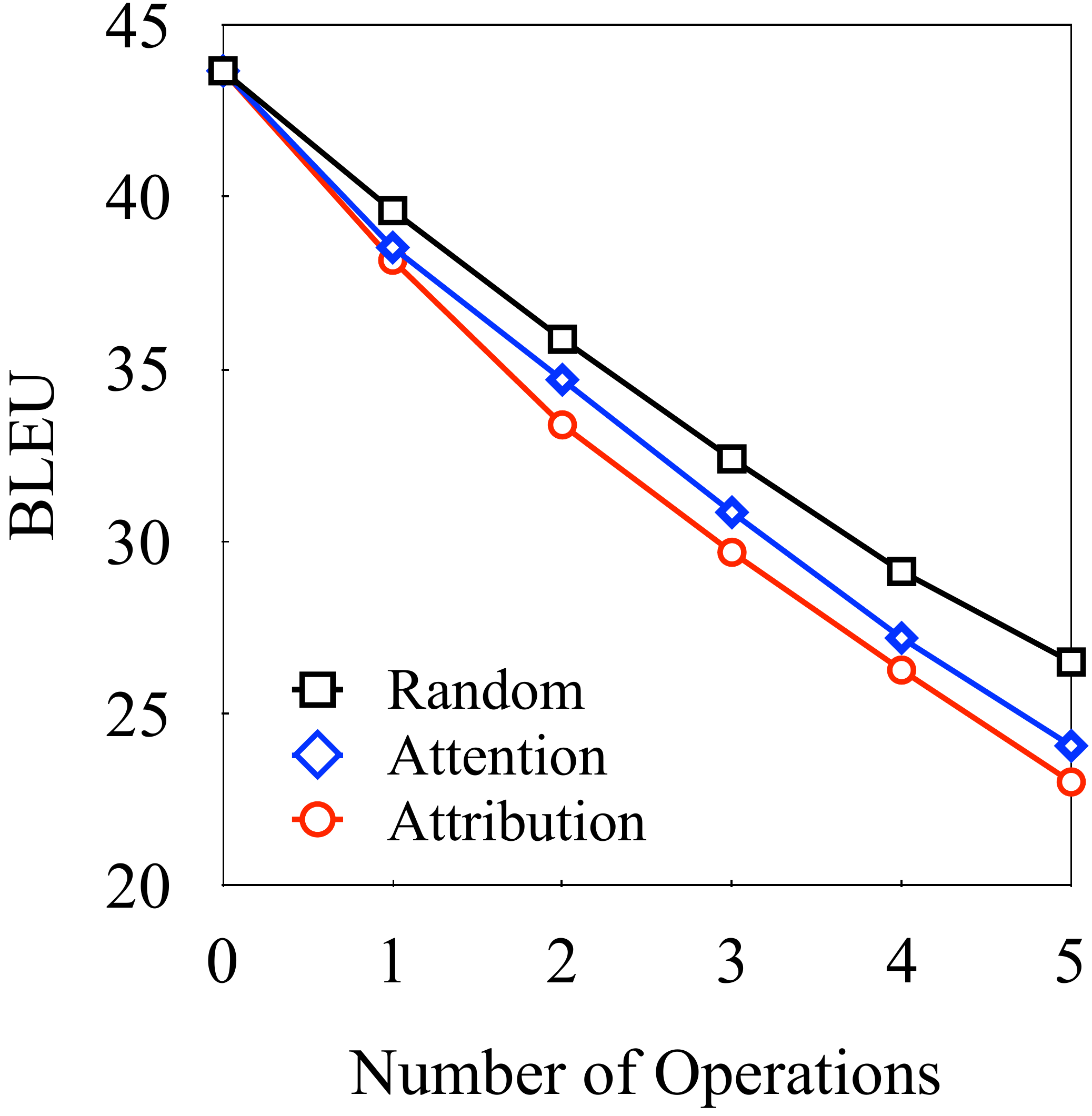}} \\
    \subfloat[\bf English$\Rightarrow$Chinese]{
    \includegraphics[width=0.3\textwidth]{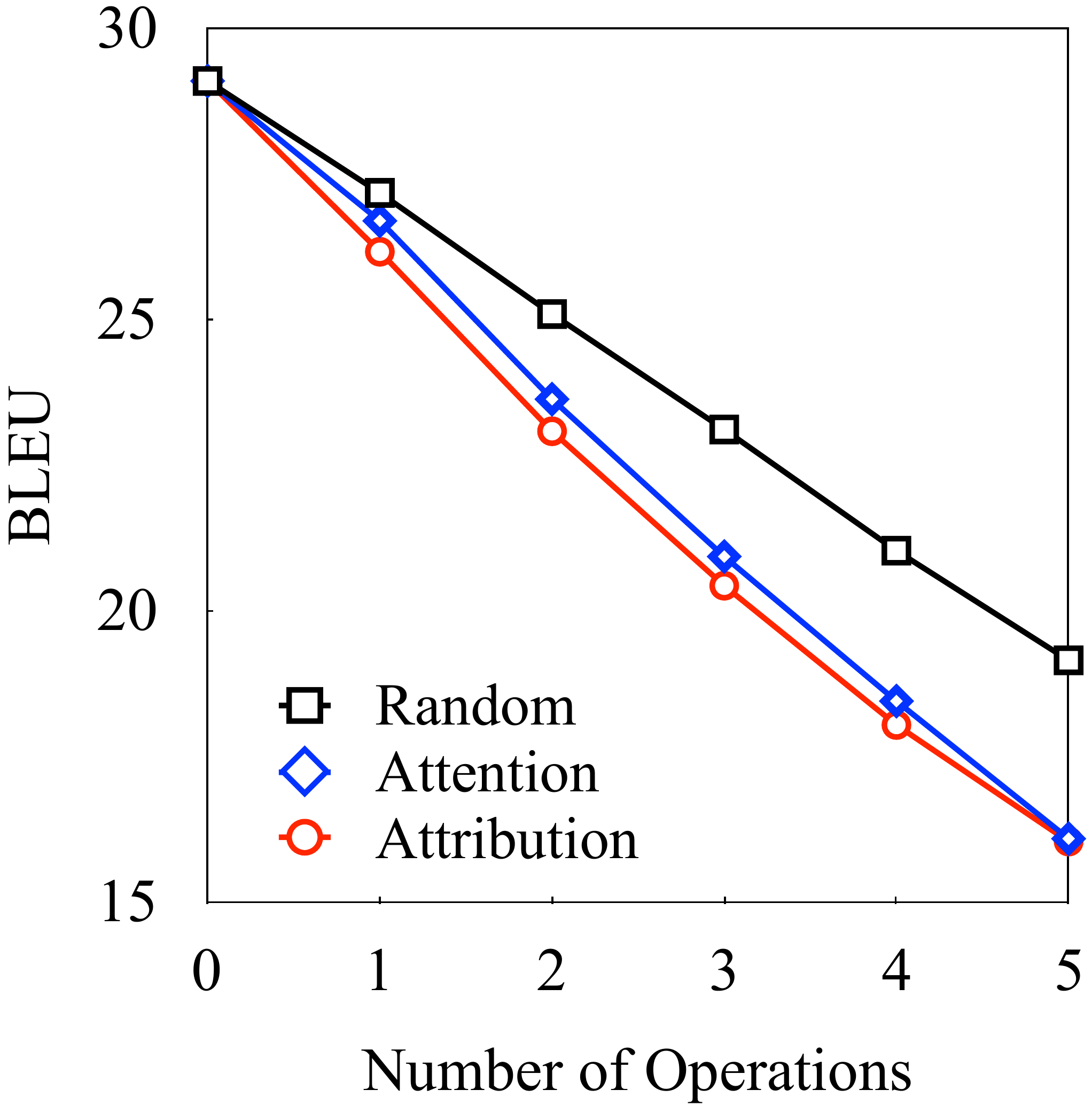}} 
    \hspace{0.05\columnwidth}
    \subfloat[\bf French$\Rightarrow$English]{
    \includegraphics[width=0.3\textwidth]{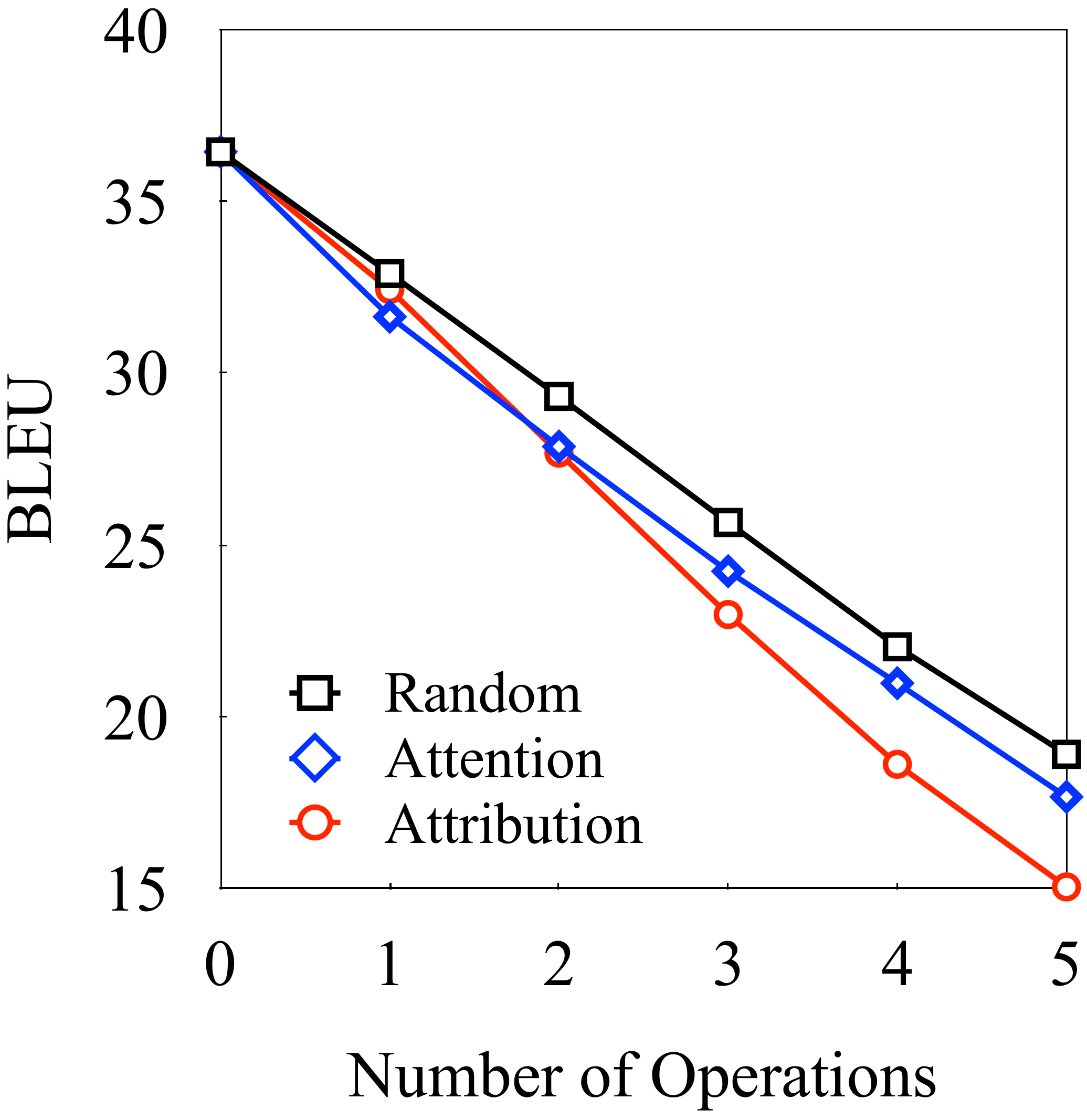}} \hspace{0.05\columnwidth}
    \subfloat[\bf Japanese$\Rightarrow$English]{
    \includegraphics[width=0.3\textwidth]{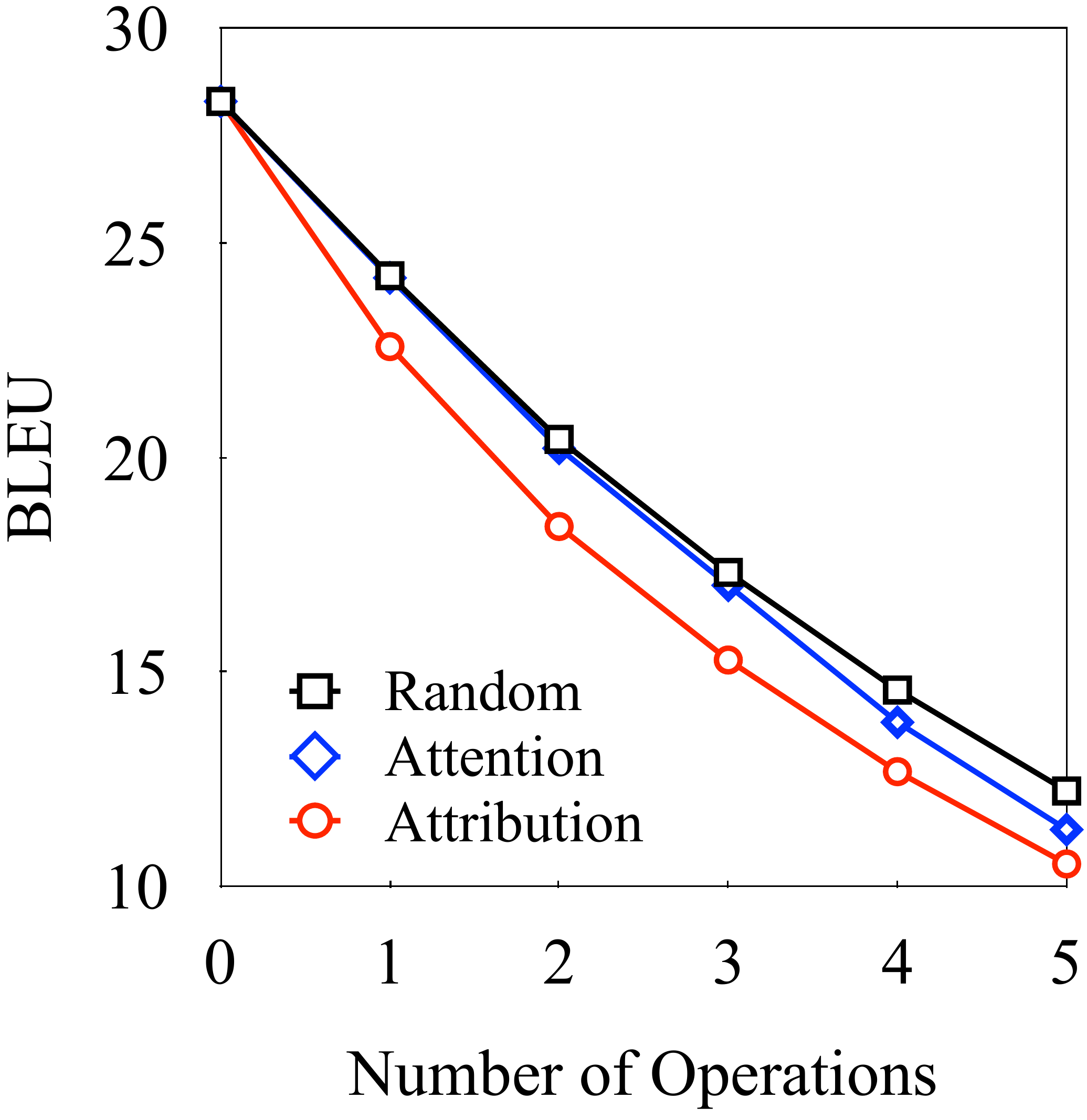}}
\caption{Effect of the {\em Mask} perturbation on (a) Chinese$\Rightarrow$English translation using the RNN-Search model, (b, c, d, e, f) other language pairs and directions using Transformer model.} 
\label{fig:model_language}
\end{figure*}

\subsection{Results on Different Perturbations}
\label{sec-exp-perturbation}

In this experiment, we investigate the effectiveness of word importance estimation methods under different synthetic perturbations. Since the perturbation on text is notoriously hard~\cite{zhang2019generating} due to the semantic shifting problem, in this experiment, we investigate three types of perturbations to avoid the potential bias :
\begin{itemize}
    \item {\em Deletion} perturbation removes the selected words from the input sentence, and it can be regarded as a specific instantiation of sentence compression~\cite{cohn2008sentence}.
    \item {\em Mask} perturbation replaces embedding vectors of the selected words with all-zero vectors~\cite{arras2016explaining}, which is similar to \textit{Deletion} perturbation except that it retains the placeholder.
    \item {\em Grammatical Replacement} perturbation replaces a word by another word of the same linguistic role (i.e., POS tags), yielding a sentence that is grammatically correct but semantically nonsensical~\cite{chomsky2002syntactic, gulordava2018colorless}, such as ``\textit{colorless green ideas sleep furiously}''.
\end{itemize}

Figure~\ref{fig:perturbations} illustrates the experimental results on Chinese$\Rightarrow$English translation with Transformer. It shows that \textit{Attribution} method consistently outperforms other methods against different perturbations on a various number of operations. Here the operation number denotes the number of perturbed words in a sentence. Specifically, we can make the following observations.

\paragraph{Important words are more influential on translation performance than the others.} Under three different perturbations, perturbing words of top-most importance leads to lower BLEU scores than \textit{Random} selected words. It confirms the existence of important words, which have greater impacts on translation performance. Furthermore, perturbing important words identified by \textit{Attribution} outperforms the \textit{Random} method by a large margin (more than 4.0 BLEU under 5 operations).

\paragraph{The gradient-based method is superior to comparative methods (e.g., Attention) in estimating word importance.}
Figure~\ref{fig:perturbations} shows that two \textit{black-box} methods (i.e., \textit{Content}, \textit{Frequency}) perform only slightly better than the \textit{Random} method. Specifically, the \textit{Frequency} method demonstrates even worse performances under the \textit{Mask} perturbation. Therefore, linguistic properties (such as POS tags) and the word frequency can only partially help identify the important words, but it is not as accurate as we thought. In the meanwhile, it is intriguing to explore what exact linguistic characteristics these important words reveal, which will be introduced in Section~\ref{Analysis}. 

We also evaluate the \textit{Attention} method, which bases on the encoder-decoder attention scores at the last layer of Transformer. Note that the \textit{Attention} method is also used to simulate the best \textit{black-box} method SOCRAT, and the results show that it is more effective than \textit{black-box} methods and the \textit{Random} baseline. Given the powerful \textit{Attention} method, \textit{Attribution} method still achieves best performances under all three perturbations. Furthermore, we find that the gap between \textit{Attribution} and \textit{Attention} is notably large (around $1.0+$ BLEU difference). \textit{Attention} method does not provide as accurate word importance as the \textit{Attribution}, which exhibits the superiority of gradient-based methods and consists with the conclusion reported in the previous study~\cite{Jain2019AttentionIN}.

In addition, as shown in Figure~\ref{fig:perturbations}, the perturbation effectiveness of \textit{Deletion}, \textit{Mask}, and \textit{Grammatical Replacement} varies from strong to weak. In the following experiments, we choose \textit{Mask} as the representative perturbation operation for its moderate perturbation performance, based on which we compare two most effective methods \textit{Attribution} and \textit{Attention}.


\subsection{Results on Different NMT Architecture and Language Pairs}
\label{sec-exp-models}

\paragraph{Different NMT Architecture} We validate the effectiveness of the proposed approach using a different NMT architecture RNN-Search on the Chinese$\Rightarrow$English translation task. 
The results are shown in Figure~\ref{fig:model_language}(a). 
We observe that the \textit{Attribution} method still outperforms both \textit{Attention} method and \textit{Random} method by a decent margin. By comparing to Transformer, the results also reveal that the RNN-Search model is less robust to these perturbations. To be specific, under the setting of five operations and \textit{Attribution} method, Transformer shows a relative decrease of $55\%$ on BLEU scores while the decline of RNN-Search model is $64\%$. 


\paragraph{Different Language Pairs and Directions}
We further conduct experiments on another two language pairs (i.e., English$\Rightarrow$French, English$\Rightarrow$Japanese in Figures~\ref{fig:model_language}(b, c)) as well as the reverse directions (Figures~\ref{fig:model_language}(d, e, f)) using Transformer under the \textit{Mask} perturbation. In all the cases, \textit{Attribution} shows the best performance while \textit{Random} achieves the worst result. More specifically, \textit{Attribution} method shows similar translation quality degradation on all three language-pairs, which declines to around the half of the original BLEU score with five operations. 


\subsection{Comparison with Supervised Erasure}
\label{sec-erasure}

\begin{figure}[t]
    \centering
    \includegraphics[width=0.3\textwidth]{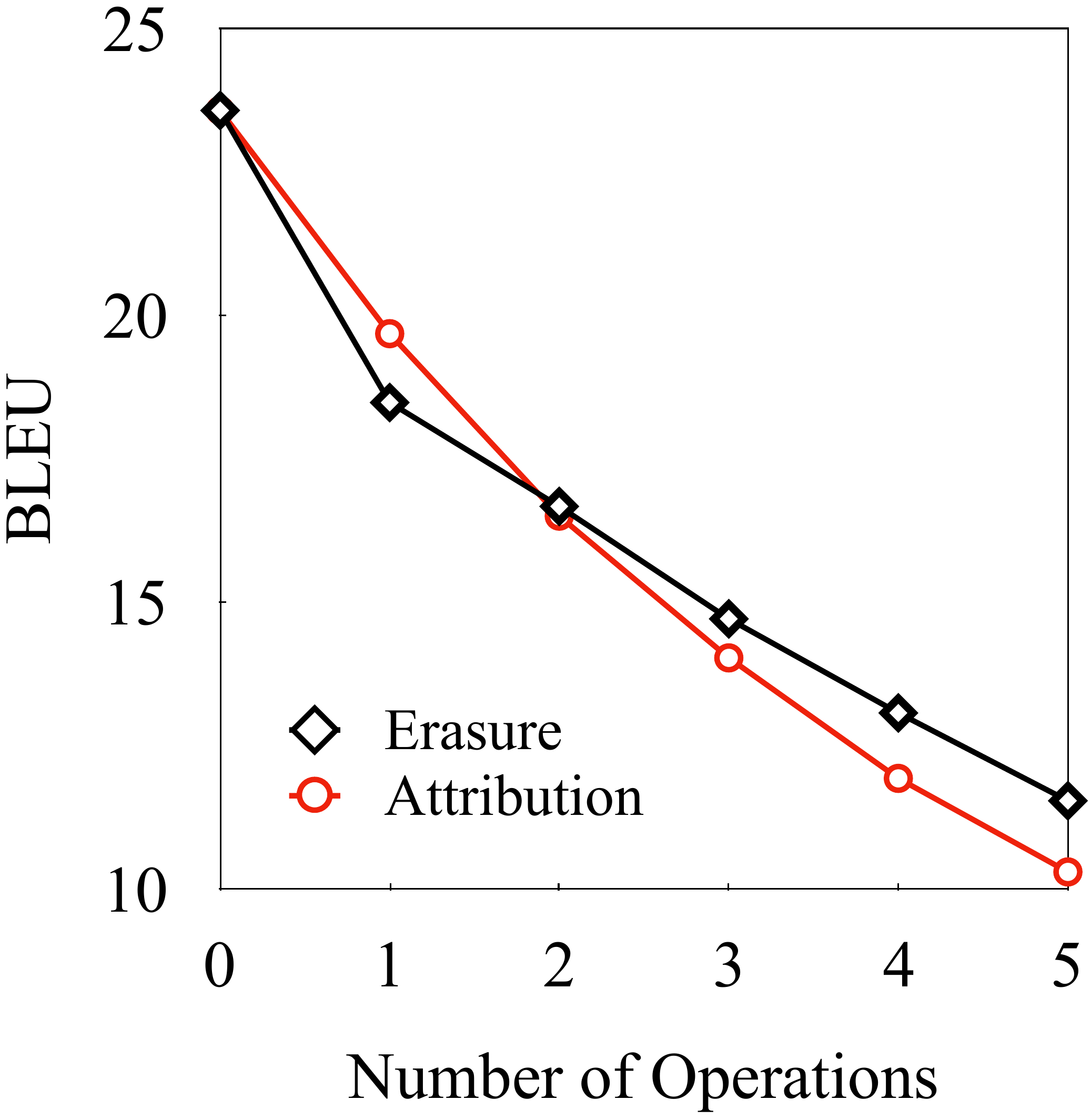}
    \caption{Effect of \textit{Attribution} and \textit{Erasure} methods on Chinese$\Rightarrow$English translation with \textit{Mask} perturbation.}
    \label{fig:erasure}
\end{figure}

There exists another straightforward method, \textit{Erasure}~\cite{alvarez2017causal, arras2016explaining, zintgraf2017visualizing}, which directly evaluates the word importance by measuring the translation performance degradation of each word. Specifically, it erases (i.e., \textit{Mask}) one word from the input sentence each time and uses the BLEU score changes to denote the word importance (after normalization). 

In Figure~\ref{fig:erasure}, we compare \textit{Erasure} method with \textit{Attribution} method under the \textit{Mask} perturbation. The results show that \textit{Attribution} method is less effective than \textit{Erasure} method when only one word is perturbed. But it outperforms the \textit{Erasure} method when perturbing 2 or more words. The results reveal that the importance calculated by erasing only one word cannot be generalized to multiple-words scenarios very well. Besides, the \textit{Erasure} method is a supervised method which requires ground-truth references, and finding a better words combination is computation infeasible when erasing multiple words.

We close this section by pointing out that our gradient-based method consistently outperforms its black-box counterparts in various settings, demonstrating the effectiveness and universality of exploiting gradients for estimating word importance. In addition, our approach is on par with or even outperforms the supervised erasure method (on multiple-word perturbations). This is encouraging since our approach does not require any external resource and is fully unsupervised.


\section{Analysis} 
\label{Analysis}

\begin{table}[t]
\centering
\begin{tabular}{c|c|c|c}
\bf Method & Top 5\% & Top 10\% & Top 15\% \\
\hline \hline
Attention  & 0.058 &0.077& 0.119\\
Erasure     &0.154  &0.170 &0.192 \\
\hline
Attribution   & \bf 0.248 & \bf 0.316  & \bf 0.342\\
\end{tabular}
\caption{\label{table:under_trans} F1 accuracy of detecting under-translation errors with the estimated word importance.}
\end{table}

In this section, we conduct analyses on two potential usages of word importance, which can help debug NMT models (Section~\ref{sec-detec-error}) and design better architectures for specific languages (Section~\ref{sec-ling-analysis}).
Due to the space limitation, we only analyze the results of Chinese$\Rightarrow$English, English$\Rightarrow$French, and English$\Rightarrow$Japanese. We list the results on the reverse directions in Appendix, in which the general conclusions also hold.

\subsection{Effect on Detecting Translation Errors}
\label{sec-detec-error}

In this experiment, we propose to use the estimated word importance to detect the under-translated words by NMT models. Intuitively, under-translated input words should contribute little to the NMT outputs, yielding much smaller word importance. Given 500 Chinese$\Rightarrow$English sentence pairs translated by the Transformer model (BLEU 23.57), we ask ten human annotators to manually label the under-translated input words, and at least two annotators label each input-hypothesis pair. These annotators have at least six years of English study experience, whose native language is Chinese. Among these sentences, 178 sentences have under-translation errors with 553 under-translated words in total.

Table~\ref{table:under_trans} lists the accuracy of detecting under-translation errors by comparing words of \textit{least} importance and human-annotated under-translated words. As seen, our \textit{Attribution} method consistently and significantly outperforms both \textit{Erasure} and \textit{Attention} approaches. 
By exploiting the word importance calculated by \textit{Attribution} method, we can identify the under-translation errors automatically without the involvement of human interpreters. Although the accuracy is not high, it is worth noting that our under-translation method is very simple and straightforward. This is potentially useful for debugging NMT models, e.g., automatic post-editing with constraint decoding~\cite{hokamp2017lexically, post2018fast}.

\begin{table}[t]
\centering
\begin{tabular}{c|c||r|r|r}
\multicolumn{2}{c||}{\bf Type}   & {\bf Zh$\Rightarrow$En} & {\bf En$\Rightarrow$Fr} &{\bf En$\Rightarrow$Ja}\\
\hline \hline
\multirow{7}{*}{\rotatebox[origin=c]{90}{{\bf POS Tags}}}
&   Noun    &\textbf{21.0\%}	 &1.9\%	 &0.7\%\\
&   Verb    &0.3\%	 &\textbf{25.0\%}	 &0.3\%\\
&   Adj.    &0.4\%	 &9.3\%	 &0.7\%\\
&   Prep.   &1.3\%	 &4.5\%	 &\textbf{26.7\%}\\
&   Dete.   &3.0\%	 &5.7\%	 &2.1\%\\
&   Punc.   &3.5\%	 &\textbf{18.3\%}	 &\textbf{30.5\%}\\
&   Others  &0.5\%	 &1.2\%	 &4.7\%\\
\hline
\multirow{4}{*}{\rotatebox[origin=c]{90}{{\bf Fertility}}}
&   $\ge 2$ &\textbf{50.2\%}	 &\textbf{21.4\%}	 &\textbf{21.7\%}\\
&   $1$     &\textbf{15.4\%}	 &7.0\%	 &3.1\%\\
&   $(0,1)$ &2.5\%	 &0.4\%	 &3.0\%\\
&   $0$     &0.0\%	 &1.9\%	 &3.8\%\\
\hline
\multirow{3}{*}{\rotatebox[origin=c]{90}{{\bf Syntactic}}}
&   Low     &1.6\%	 &2.5\%	 &1.2\%	 \\
&   Middle  &0.3\%	 &0.8\%	 &1.4\%     \\
&   High    &0.0\%	 &0.1\%	 &0.1\%	 \\
\end{tabular}
\caption{\label{table:decision_tree_result} Correlation between \textit{Attribution} word importance with POS tags, Fertility, and Syntactic Depth. Fertility can be categorized into 4 types: one-to-many (``$\ge 2$''), one-to-one (``$1$''), many-to-one (``$(0,1)$''), and null-aligned (``$0$''). Syntactic depth shows the depth of a word in the dependency tree. 
A lower tree depth indicates closer to the root node in the dependency tree, which might indicate a more important word.}
\end{table}

\begin{table*}[t] 
\centering
\begin{tabular}{c|c||rr|r||rr|r||rr|r}
\multicolumn{2}{c||}{\bf Type}   &  \multicolumn{3}{c||}{\bf Chinese$\Rightarrow$English} & \multicolumn{3}{c||}{\bf English$\Rightarrow$French} & \multicolumn{3}{c}{\bf English$\Rightarrow$Japanese}\\
\cline{3-11} 
\multicolumn{2}{c||}{}   &   Count   &    Attri. &   $\bm \bigtriangleup$   &   Count   &    Attri. &   $\bm \bigtriangleup$    &    Count   &    Attri.  &   $\bm \bigtriangleup$  \\
\hline \hline
\multirow{4}{*}{\rotatebox[origin=c]{90}{{\bf Content}}}
&   Noun    &0.383	&0.407	&\em+6.27\%	&0.341	&0.355	&\em +4.11\%	&0.365	&0.336	&\em-7.95\% \\
&   Verb    &0.165	&0.160	&\em-3.03\%	&0.146	&0.131	&\em -10.27\%	&0.127	&0.123	&\em-3.15\%  \\
&   Adj.    &0.032	&0.029	&\em-9.38\%	&0.076	&0.072	&\em -5.26\%	&0.094	&0.088	&\em-6.38\% \\
\cdashline{2-11}
&   Total   &0.579	&0.595	&\em+2.76\%	&0.563	&0.558	&\em -0.89\%	&0.587	&0.547	&\em-6.81\% \\
\hline
\multirow{5}{*}{\rotatebox[origin=c]{90}{{\bf Content-Free}}} 
&   Prep.   &0.056	&0.051	&\em-8.93\%	&0.120	&0.132	&\em +10.00\%	&0.129	&0.151	&\em+17.05\% \\
&   Dete.   &0.043	&0.043	&\em0.00\%	&0.102	&0.101	&\em -0.98\%	&0.112	&0.103	&\em-8.04\% \\
&   Punc.   &0.137	&0.131	&\em-4.38\%	&0.100	&0.091	&\em -9.00\%	&0.096	&0.120	&\em+25.47\% \\
&   Others  &0.186	&0.179	&\em-3.76\%	&0.115	&0.118	&\em +2.61\%	&0.076	&0.079	&\em+3.95\% \\
\cdashline{2-11}
&   Total   &0.421	&0.405	&\em-3.80\%	&0.437	&0.442	&\em +1.14\%	&0.413	&0.453	&\em+9.69\%  \\
\end{tabular}
\caption{\label{table:syntactic_distribution} Distribution of syntactic categories (e.g. content words vs. content-free words) based on word count (``Count'') and \textit{Attribution} importance (``Attri.''). ``$\bigtriangleup$'' denotes relative change over the count-based distribution.}
\end{table*}

\begin{table*}[t]
\centering
\begin{tabular}{c||rr|r||rr|r||rr|r}
\multirow{2}{*}{\bf Fertility}   &  \multicolumn{3}{c||}{\bf Chinese$\Rightarrow$English} & \multicolumn{3}{c||}{\bf English$\Rightarrow$French} & \multicolumn{3}{c}{\bf English$\Rightarrow$Japanese}\\
\cline{2-10}
   &   Count   &    Attri. &   $\bm \bigtriangleup$   &   Count   &    Attri. &   $\bm \bigtriangleup$    &    Count   &    Attri.  &   $\bm \bigtriangleup$  \\
\hline \hline
$\ge 2$    &0.087	&0.146	&\em +67.82\%	&0.126	&0.138	&\em +9.52\%	&0.117	&0.143	&\em +22.22\% \\
\hline
$1$        &0.621	&0.622	&\em +0.16\%	&0.672	&0.670	&\em -0.30\%	&0.570	&0.565	&\em -0.88\% \\
$(0, 1)$   &0.115	&0.081	&\em -29.57\%	&0.116	&0.113	&\em -2.59\%	&0.059	&0.055	&\em -6.78\% \\
\hline
$0$        &0.176	&0.150	&\em -14.77\%	&0.086	&0.079	&\em -8.14\%	&0.254	&0.237	&\em -6.69\% \\
\end{tabular}
\caption{\label{table:fertility_distribution} Distributions of word fertility and their relative change based on \textit{Attribution} importance and word count.}
\end{table*}

\subsection{Analysis on Linguistic Properties}
\label{sec-ling-analysis}

In this section, we analyze the linguistic characteristics of important words identified by the attribution-based approach. Specifically, we investigate several representative sets of linguistic properties, including POS tags, and fertility, and depth in a syntactic parse tree.  In these analyses, we multiply the word importance with the corresponding sentence length for fair comparison.
We use a decision tree based regression model to calculate the correlation between the importance and linguistic properties. 

Table~\ref{table:decision_tree_result} lists the correlations, where a higher value indicates a stronger correlation. We find that the syntactic information is almost independent of the word importance value. Instead, the word importance strongly correlates with the POS tags and fertility features, and these features in total contribute over 95\%.
Therefore, in the following analyses, we mainly focus on the POS tags (Table~\ref{table:syntactic_distribution}) and fertility properties (Table~\ref{table:fertility_distribution}). 
For better illustration, we calculate the distribution over the linguistic property based on both the \textit{Attribution} importance (``Attr.'') and the word frequency (``Count'') inside a sentence. The larger the relative increase between these two values, the more important the linguistic property is.

\paragraph{Certain syntactic categories have higher importance while the categories vary across language pairs.} As shown in Table~\ref{table:syntactic_distribution}, \textit{content} words are more important on Chinese$\Rightarrow$English but \textit{content-free} words are more important on English$\Rightarrow$Japanese. On English$\Rightarrow$French, there is no notable increase or decrease of the distribution since English and French are in essence very similar. 
We also obtain some specific findings of great interest. For example, we find that noun is more important on Chinese$\Rightarrow$English translation, while preposition is more important on English$\Rightarrow$French translation. More interestingly, English$\Rightarrow$Japanese translation shows a substantial discrepancy in contrast to the other two language pairs. The results reveal that preposition and punctuation are very important in English$\Rightarrow$Japanese translation, which is counter-intuitive.

Punctuation in NMT is understudied since it carries little information and often does not affect the understanding of a sentence. However, we find that punctuation is important on English$\Rightarrow$Japanese translation, whose proportion increases dramatically. We conjecture that it is because the punctuation could affect the sense groups in a sentence, which further benefits the syntactic reordering in Japanese.

\paragraph{Words of high fertility are always important.}  
We further compare the fertility distribution based on word importance and the word frequency on three language pairs. We hypothesize that a source word that corresponds to multiple target words should be more important since it contributes more to both sentence length and BLEU score. 

Table~\ref{table:fertility_distribution} lists the results. Overall speaking, one-to-many fertility is consistently more important on all three language pairs, which confirms our hypothesis. On the contrary, null-aligned words receive much less attention, which shows a persistently decrease on three language pairs. It is also reasonable since null-aligned input words contribute almost nothing to the translation outputs.

\section{Discussion and Conclusion}

We approach understanding NMT by investigating the word importance via a gradient-based method, which bridges the gap between word importance and translation performance.
Empirical results show that the gradient-based method is superior to several black-box methods in estimating the word importance. Further analyses show that important words are of distinct syntactic categories on different language pairs, which might support the viewpoint that essential inductive bias should be introduced into the model design~\cite{strubell2018linguistically}. Our study also suggests the possibility of detecting the notorious under-translation problem via the gradient-based method.

This paper is an initiating step towards the general understanding of NMT models, which may bring some potential improvements, such as
\begin{itemize}
    \item {\em Interactive MT and Constraint Decoding} ~\cite{Foster:1997:MT,hokamp2017lexically}: The model pays more attention to the detected unimportant words, which are possibly under-translated;
    \item {\em Adaptive Input Embedding} \cite{Baevski:2019:ICLR}: We can extend the adaptive softmax~\cite{Grave:2017:ICML} to the input embedding of variable capacity -- more important words are assigned with more capacity; 
    \item {\em NMT Architecture Design}: The language-specific inductive bias (e.g., different behaviors on POS) should be incorporated into the model design.
\end{itemize}
We can also explore other applications of word importance to improve NMT models, such as more tailored training methods. 
In general, model interpretability can build trust in model predictions, help error diagnosis and facilitate model refinement.
We expect our work could shed light on the NMT model understanding and benefit the model improvement.

There are many possible ways to implement the general idea of exploiting gradients for model interpretation. The aim of this paper is not to explore this whole space but simply to show that some fairly straightforward implementations work well. 
Our approach can benefit from advanced exploitation of the gradients or other useful intermediate information, which we leave to the future work.

\section*{Acknowledgement}

Shilin He and Michael R. Lyu were supported by the Research Grants Council of the Hong Kong Special Administrative Region, China (No. CUHK 14210717 of the General Research Fund), and Microsoft Research Asia (2018 Microsoft Research Asia Collaborative Research Award). We thank the anonymous reviewers for their insightful comments and suggestions.

\balance
\bibliography{emnlp-ijcnlp-2019}

\begin{thebibliography}{32}
\expandafter\ifx\csname natexlab\endcsname\relax\def\natexlab#1{#1}\fi

\bibitem[{Alvarez-Melis and Jaakkola(2017)}]{alvarez2017causal}
David Alvarez-Melis and Tommi Jaakkola. 2017.
\newblock A causal framework for explaining the predictions of black-box
  sequence-to-sequence models.
\newblock In \emph{EMNLP}.

\bibitem[{Arras et~al.(2016)Arras, Horn, Montavon, M{\"u}ller, and
  Samek}]{arras2016explaining}
Leila Arras, Franziska Horn, Gr{\'e}goire Montavon, Klaus-Robert M{\"u}ller,
  and Wojciech Samek. 2016.
\newblock Explaining predictions of non-linear classifiers in nlp.
\newblock In \emph{Proceedings of the 1st Workshop on Representation Learning
  for NLP}.

\bibitem[{Baevski and Auli(2019)}]{Baevski:2019:ICLR}
Alexei Baevski and Michael Auli. 2019.
\newblock Adaptive input representations for neural language modeling.
\newblock In \emph{ICLR}.

\bibitem[{Bahdanau et~al.(2014)Bahdanau, Cho, and Bengio}]{bahdanau2014neural}
Dzmitry Bahdanau, Kyunghyun Cho, and Yoshua Bengio. 2014.
\newblock Neural machine translation by jointly learning to align and
  translate.
\newblock In \emph{ICLR}.

\bibitem[{Bau et~al.(2019)Bau, Belinkov, Sajjad, Durrani, Dalvi, and
  Glass}]{bau2018identifying}
Anthony Bau, Yonatan Belinkov, Hassan Sajjad, Nadir Durrani, Fahim Dalvi, and
  James Glass. 2019.
\newblock Identifying and controlling important neurons in neural machine
  translation.
\newblock In \emph{ICLR}.

\bibitem[{Belinkov et~al.(2017)Belinkov, Durrani, Dalvi, Sajjad, and
  Glass}]{Belinkov:2017:ACL}
Yonatan Belinkov, Nadir Durrani, Fahim Dalvi, Hassan Sajjad, and James Glass.
  2017.
\newblock {What do neural machine translation models learn about morphology?}
\newblock In \emph{ACL}.

\bibitem[{Chomsky and Lightfoot(2002)}]{chomsky2002syntactic}
Noam Chomsky and David~W Lightfoot. 2002.
\newblock \emph{Syntactic structures}.
\newblock Walter de Gruyter.

\bibitem[{Cohn and Lapata(2008)}]{cohn2008sentence}
Trevor Cohn and Mirella Lapata. 2008.
\newblock Sentence compression beyond word deletion.
\newblock In \emph{COLING}.

\bibitem[{Dhamdhere et~al.(2019)Dhamdhere, Sundararajan, and
  Yan}]{dhamdhere2018important}
Kedar Dhamdhere, Mukund Sundararajan, and Qiqi Yan. 2019.
\newblock How important is a neuron?
\newblock In \emph{ICLR}.

\bibitem[{Ding et~al.(2017)Ding, Liu, Luan, and Sun}]{ding2017visualizing}
Yanzhuo Ding, Yang Liu, Huanbo Luan, and Maosong Sun. 2017.
\newblock Visualizing and understanding neural machine translation.
\newblock In \emph{ACL}.

\bibitem[{Foster et~al.(1997)Foster, Isabelle, and Plamondon}]{Foster:1997:MT}
George Foster, Pierre Isabelle, and Pierre Plamondon. 1997.
\newblock Target-text mediated interactive machine translation.
\newblock \emph{Machine Translation}, 12(1/2):175--194.

\bibitem[{Gehring et~al.(2017)Gehring, Auli, Grangier, Yarats, and
  Dauphin}]{gehring2017convolutional}
Jonas Gehring, Michael Auli, David Grangier, Denis Yarats, and Yann~N Dauphin.
  2017.
\newblock Convolutional sequence to sequence learning.
\newblock In \emph{ICML}.

\bibitem[{Grave et~al.(2017)Grave, Joulin, Ciss{\'e}, Grangier, and
  J{\'e}gou}]{Grave:2017:ICML}
{\'E}douard Grave, Armand Joulin, Moustapha Ciss{\'e}, David Grangier, and
  Herv{\'e} J{\'e}gou. 2017.
\newblock Efficient softmax approximation for {GPU}s.
\newblock In \emph{ICML}.

\bibitem[{Gulordava et~al.(2018)Gulordava, Bojanowski, Grave, Linzen, and
  Baroni}]{gulordava2018colorless}
Kristina Gulordava, Piotr Bojanowski, Edouard Grave, Tal Linzen, and Marco
  Baroni. 2018.
\newblock Colorless green recurrent networks dream hierarchically.
\newblock In \emph{NAACL}.

\bibitem[{Hassan et~al.(2018)Hassan, Aue, Chen, Chowdhary, Clark, Federmann,
  Huang, Junczys-Dowmunt, Lewis, Li et~al.}]{hassan2018achieving}
Hany Hassan, Anthony Aue, Chang Chen, Vishal Chowdhary, Jonathan Clark,
  Christian Federmann, Xuedong Huang, Marcin Junczys-Dowmunt, William Lewis,
  Mu~Li, et~al. 2018.
\newblock Achieving human parity on automatic chinese to english news
  translation.
\newblock In \emph{arXiv:1803.05567}.

\bibitem[{Hokamp and Liu(2017)}]{hokamp2017lexically}
Chris Hokamp and Qun Liu. 2017.
\newblock Lexically constrained decoding for sequence generation using grid
  beam search.
\newblock In \emph{ACL}.

\bibitem[{Jain and Wallace(2019)}]{Jain2019AttentionIN}
Sarthak Jain and Byron~C. Wallace. 2019.
\newblock Attention is not explanation.
\newblock In \emph{NAACL}.

\bibitem[{L{\"a}ubli et~al.(2018)L{\"a}ubli, Sennrich, and
  Volk}]{Laubli:2018:EMNLP}
Samuel L{\"a}ubli, Rico Sennrich, and Martin Volk. 2018.
\newblock {Has Machine Translation Achieved Human Parity? A Case for
  Document-level Evaluation}.
\newblock In \emph{EMNLP}.

\bibitem[{Li et~al.(2016)Li, Monroe, and Jurafsky}]{li2016understanding}
Jiwei Li, Will Monroe, and Dan Jurafsky. 2016.
\newblock Understanding neural networks through representation erasure.
\newblock In \emph{arXiv preprint arXiv:1612.08220}.

\bibitem[{Morishita et~al.(2017)Morishita, Suzuki, and
  Nagata}]{morishita2017ntt}
Makoto Morishita, Jun Suzuki, and Masaaki Nagata. 2017.
\newblock Ntt neural machine translation systems at wat 2017.
\newblock In \emph{WAT}.

\bibitem[{Mudrakarta et~al.(2018)Mudrakarta, Taly, Sundararajan, and
  Dhamdhere}]{mudrakarta2018did}
Pramod~Kaushik Mudrakarta, Ankur Taly, Mukund Sundararajan, and Kedar
  Dhamdhere. 2018.
\newblock Did the model understand the question?
\newblock In \emph{ACL}.

\bibitem[{Post and Vilar(2018)}]{post2018fast}
Matt Post and David Vilar. 2018.
\newblock Fast lexically constrained decoding with dynamic beam allocation for
  neural machine translation.
\newblock In \emph{NAACL}.

\bibitem[{Schwarzenberg et~al.(2019)Schwarzenberg, Harbecke, Macketanz,
  Avramidis, and M{\"o}ller}]{schwarzenberg2019train}
Robert Schwarzenberg, David Harbecke, Vivien Macketanz, Eleftherios Avramidis,
  and Sebastian M{\"o}ller. 2019.
\newblock Train, sort, explain: Learning to diagnose translation models.
\newblock In \emph{NAACL}.

\bibitem[{Selvaraju et~al.(2017)Selvaraju, Cogswell, Das, Vedantam, Parikh, and
  Batra}]{selvaraju2017grad}
Ramprasaath~R Selvaraju, Michael Cogswell, Abhishek Das, Ramakrishna Vedantam,
  Devi Parikh, and Dhruv Batra. 2017.
\newblock Grad-cam: Visual explanations from deep networks via gradient-based
  localization.
\newblock In \emph{ICCV}.

\bibitem[{Sennrich et~al.(2016)Sennrich, Haddow, and
  Birch}]{sennrich2015neural}
Rico Sennrich, Barry Haddow, and Alexandra Birch. 2016.
\newblock Neural machine translation of rare words with subword units.
\newblock In \emph{ACL}.

\bibitem[{Shi et~al.(2016)Shi, Padhi, and Knight}]{Shi:2016:EMNLP}
Xing Shi, Inkit Padhi, and Kevin Knight. 2016.
\newblock {Does string-based neural mt learn source syntax?}
\newblock In \emph{EMNLP}.

\bibitem[{Strubell et~al.(2018)Strubell, Verga, Andor, Weiss, and
  McCallum}]{strubell2018linguistically}
Emma Strubell, Patrick Verga, Daniel Andor, David Weiss, and Andrew McCallum.
  2018.
\newblock {Linguistically-Informed Self-Attention for Semantic Role Labeling}.
\newblock In \emph{EMNLP}.

\bibitem[{Sundararajan et~al.(2017)Sundararajan, Taly, and
  Yan}]{sundararajan2017axiomatic}
Mukund Sundararajan, Ankur Taly, and Qiqi Yan. 2017.
\newblock Axiomatic attribution for deep networks.
\newblock In \emph{ICML}.

\bibitem[{Vaswani et~al.(2017)Vaswani, Shazeer, Parmar, Uszkoreit, Jones,
  Gomez, Kaiser, and Polosukhin}]{vaswani2017attention}
Ashish Vaswani, Noam Shazeer, Niki Parmar, Jakob Uszkoreit, Llion Jones,
  Aidan~N Gomez, {\L}ukasz Kaiser, and Illia Polosukhin. 2017.
\newblock Attention is all you need.
\newblock In \emph{NeurIPS}.

\bibitem[{Yang et~al.(2019)Yang, Wang, Wong, Chao, and Tu}]{Yang:2019:ACL}
Baosong Yang, Longyue Wang, Derek~F. Wong, Lidia~S. Chao, and Zhaopeng Tu.
  2019.
\newblock Assessing the ability of self-attention networks to learn word order.
\newblock In \emph{ACL}.

\bibitem[{Zhang et~al.(2019)Zhang, Sheng, and Alhazmi}]{zhang2019generating}
Wei~Emma Zhang, Quan~Z Sheng, and Ahoud Abdulrahmn~F Alhazmi. 2019.
\newblock Generating textual adversarial examples for deep learning models: A
  survey.
\newblock In \emph{arXiv preprint arXiv:1901.06796}.

\bibitem[{Zintgraf et~al.(2017)Zintgraf, Cohen, Adel, and
  Welling}]{zintgraf2017visualizing}
Luisa~M Zintgraf, Taco~S Cohen, Tameem Adel, and Max Welling. 2017.
\newblock Visualizing deep neural network decisions: Prediction difference
  analysis.
\newblock In \emph{ICLR}.

\end{thebibliography}
\bibliographystyle{acl_natbib}

\pagebreak

\appendix 

\onecolumn
\section{Analyses on Reverse Directions}

\begin{table}[h] 
\centering
\begin{tabular}{c|c||rr|r||rr|r||rr|r}
\multicolumn{2}{c||}{\bf Type}   &  \multicolumn{3}{c||}{\bf English$\Rightarrow$Chinese} & \multicolumn{3}{c||}{\bf French$\Rightarrow$English} & \multicolumn{3}{c}{\bf Japanese$\Rightarrow$English}\\
\cline{3-11} 
\multicolumn{2}{c||}{}   &   Count   &    Attri. &   $\bm \bigtriangleup$   &   Count   &    Attri. &   $\bm \bigtriangleup$    &    Count   &    Attri.  &   $\bm \bigtriangleup$  \\
\hline \hline
\multirow{4}{*}{\rotatebox[origin=c]{90}{{\bf Content}}}
&   Noun    &0.313	&0.338	&\em +7.99\%	&0.323	&0.313	&\em -3.10\%	&0.426	&0.377	&\em -11.50\%\\
&   Verb    &0.132	&0.127	&\em -3.79\%	&0.172	&0.160	&\em -6.98\%	&0.091	&0.085	&\em -6.59\%\\
&   Adj.    &0.091	&0.094	&\em +3.30\%	&0.078	&0.077	&\em -1.28\%	&0.014	&0.012	&\em -14.29\%\\
\cdashline{2-11}
&   Total   &0.536	&0.559	&\em +4.29\%	&0.572	&0.551	&\em -3.67\%	&0.531	&0.473	&\em -10.92\%\\
\hline
\multirow{5}{*}{\rotatebox[origin=c]{90}{{\bf Content-Free}}} 
&   Prep.   &0.133	&0.129	&\em -3.01\%	&0.116	&0.125	&\em +7.76\%		&-	&-	&-\\
&   Dete.   &0.122	&0.113	&\em -7.38\%	&0.123	&0.126	&\em +2.44\%		&-	&-	&-\\
&   Punc.   &0.088	&0.078	&\em -11.36\%	&0.076	&0.084	&\em +10.53\%	&0.091	&0.122	&\em +34.07\%\\
&   Others  &0.121	&0.121	&\em 0.00\%		&0.113	&0.114	&\em +0.88\%	&0.377	&0.405	&\em +7.43\%\\
\cdashline{2-11}
&   Total   &0.464	&0.441	&\em -4.96\%	&0.428	&0.449	&\em +4.91\%	&0.469	&0.527	&\em +12.37\%\\
\end{tabular}
\caption{\label{table:reverse_syntactic_distribution} Distribution of syntactic categories with reverse directions based on word count (``Count'') and \textit{Attribution} importance (``Attri.''). ``$\bigtriangleup$'' denotes relative change over the count-based distribution.}
\end{table}

\begin{table*}[h]
\centering

\begin{tabular}{c||rr|r||rr|r||rr|r}
\multirow{2}{*}{\bf Fertility}   &  \multicolumn{3}{c||}{\bf English$\Rightarrow$Chinese} & \multicolumn{3}{c||}{\bf French$\Rightarrow$English} & \multicolumn{3}{c}{\bf Japanese$\Rightarrow$English}\\
\cline{2-10}
   &   Count   &    Attri. &   $\bm \bigtriangleup$   &   Count   &    Attri. &   $\bm \bigtriangleup$    &    Count   &    Attri.  &   $\bm \bigtriangleup$  \\
\hline \hline
$\ge 2$     &0.091	&0.106	&\em +16.48\%	&0.088	&0.094	&\em +6.82\% &0.079	&0.085	&\em+7.59\%\\
\hline
$1$         &0.616	&0.629	&\em +2.11\%	&0.707	&0.721	&\em +1.98\% &0.513	&0.520	&\em+1.36\%\\
$(0, 1)$    &0.083	&0.077	&\em -7.23\%	&0.102	&0.094	&\em -7.84\% &0.086	&0.097	&\em+12.79\%\\
\hline
$0$         &0.210	&0.187	&\em -10.95\%	&0.103	&0.092	&\em -10.68\% &0.322 &0.298	&\em-7.45\%\\
\end{tabular}
\caption{\label{table:reverse_fertility_distribution} Distributions of word fertility and relative changes with reverse directions.}
\end{table*}


\begin{multicols}{2}

We analyze the distribution of syntactic categories and word fertility on the same language pairs with reverse directions, i.e., English$\Rightarrow$Chinese, French$\Rightarrow$English, and Japanese$\Rightarrow$English. The results are shown in Table~\ref{table:reverse_syntactic_distribution} and Table~\ref{table:reverse_fertility_distribution} respectively, where we observe similar findings as before. We use the Stanford POS tagger to parse the English and French input sentences, and use the Kytea\footnote{http://www.phontron.com/kytea/} to parse the Japanese input sentences. 

\paragraph{Syntactic Categories} On English$\Rightarrow$Chinese, \textit{content} words are more important than \textit{content-free} words, while the situation is reversed on both French$\Rightarrow$English and Japanese$\Rightarrow$English translations. Since there is no clear boundary between Preposition/Determiner and other categories in Japanese, we set both categories to be none. Similarly, Punctuation is more important on Japanese$\Rightarrow$English, which is in line with the finding on English$\Rightarrow$Japanese. Overall speaking, it might indicate that the Syntactic distribution with word importance is language-pair related instead of the direction.

\paragraph{Word Fertility} The word fertility also shows similar trend as the previously reported results, where one-to-many fertility is more important and null-aligned fertility is less important. Interestingly, many-to-one fertility shows an increasing trend on Japanese$\Rightarrow$English translation, but the proportion is relatively small. 

In summary, the findings on language pairs with reverse directions still agree with the findings in the paper, which further confirms the generality of our experimental findings.

\end{multicols}

\end{document}